# A RULE-BASED APPROACH FOR ALIGNING JAPANESE-SPANISH SENTENCES FROM A COMPARABLE CORPORA

Jessica C. Ramírez[1] and Yuji Matsumoto[2]

Information Science, Nara Institute of Science and Technology, Nara, Japan

[1]Jessicrv1@yahoo.com.mx [2] matsu@is.naist.jp

## ABSTRACT

*The performance of a Statistical Machine Translation System (SMT) system is proportionally directed to the quality and length of the parallel corpus it uses. However for some pair of languages there is a considerable lack of them. The long term goal is to construct a Japanese-Spanish parallel corpus to be used for SMT, whereas, there are a lack of useful Japanese-Spanish parallel Corpus. To address this problem, In this study we proposed a method for extracting Japanese-Spanish Parallel Sentences from Wikipedia using POS tagging and Rule-Based approach. The main focus of this approach is the syntactic features of both languages. Human evaluation was performed over a sample and shows promising results, in comparison with the baseline.*

## 1. INTRODUCTION

Much research in recent years has focused on constructing semi-automatic and automatic aligned data resources, which are essential for many Natural Language Processing Tasks; however for some pairs of languages there is still a huge lack of annotated data.

Manual construction of Parallel corpus requires high quality translators, besides It is time consuming and expensive. With the proliferation of the internet and the immense amount of data, a number of researchers have proposed using the World Wide Web as a large-scale corpus[5]. However due to redundancy and ambiguous information on the web, we must find methods of extracting only the information that is useful for a given task. [4].

Extracting parallel sentences from a comparable corpus is a challenging task, due to the fact that despite two documents can be referred to a same topic, it can be possible that both documents do not have a single sentence in common.

In this study, we propose an approach for extracting Japanese-Spanish parallel sentences from Wikipedia using Part-of-Speech rule based alignment and Dictionary based translation. We use as comparable corpora to Wikipedia articles, a dictionary extracted from the Wikipedia links and Aulex a free Japanese-Spanish dictionary

## 2. RELATED WORKS

The use of Wikipedia as a data resource in NLP is fairly new, and thus research is fairly limited. There are, however, works showing promising results.[2] attempts to extract the named entities from Wikipedia and presents two disambiguation methods using cosine similarity and SVM.

Firstly detecting a named entity from Wikipedia by using IE technique, and disambiguates between multiples entities, using context articles similarity by using cosine similarity and of taxonomy kernel.

Such work that is directly related to this research is [1] . Their research uses two approaches to similarity between sentences in Wikipedia. Firstly they introduced an MT based approach, using Jaccard Similarity and 'Babel MT system of Altavista[1]' to aligned the sentences. The second approach, the Link based bilingual lexicon, that used the hyperlinks in every sentences by mean of a dictionary extracted from Wikipedia to select the. Their result showed best result on using the second approach, especially in articles that are literal translation on each other.

Our approach differs in that we convert the articles in their equivalent POS tags and we just align sentences that are according to the rules we add. And then we used two Japanese- Spanish dictionaries as a seed lexicon.

## 3. BACKGROUND

### 3.1. Comparable Corpora[2]

A comparable corpus is a collection of text about a given topic in two or more languages. For example, The 'Yomiuri Shimbun[3] corpora', a corpus extracted from their daily news both in English and Japanese. Despite the news in both languages are the same, they do not make a proper translation of the content.

Comparable corpora are used for many NLP tasks, such as: Information Retrieval, Machine Translation, bilingual lexicon extraction, and so on. In languages with a scarcity of resources Comparable corpora are an alternative of prime order in NLP research.

### 3.2. Wikipedia

Wikipedia[4] is a multilingual web-based encyclopedia with articles on a wide range of topics, in which the texts are aligned across different languages.

Wikipedia is the successor of Nupedia- an online encyclopedia written by experts in different fields that does not exist now. Wikipedia arose as a single language project (English) on January, 2001 to support Wikipedia differs from Nupedia mainly that anyone can write it, and writers do not need to be an expert in the field that is written.

---

[1] Babel, is an online multilingual  translation system
[2] Corpora is a plural of corpus.
[3] 読売新聞: is a Japanese newspaper, with an English version too.
http://www.yomiuri.co.jp/
[4] http://en.wikipedia.org/wiki/Main_Page

Wikipedia is written collaboratively by volunteers (called "wikipedians") from different places all around the world. This is because Wikipedia has volunteers from many nationalities who write in many different languages. Actually it has articles written in more than 200 languages, with different numbers of articles in each language.

The topics vary from science, covering many different fields such as informatics, biology, anthropology and entertainment, such as albums name, artist, actors etc. and fictional characters such as James Bond.

Wikipedia is not only a simple encyclopaedia; it has some features that make Wikipedia suitable for NLP research. These features are:

### 3.2.1. Redirect pages

The redirect page is a very suitable resource for eliminating redundant articles. This means avoiding the existence of two articles referring to the same topic.

This is used in the case of:

- Synonyms like 'altruism' which redirects to 'selflessness'
- Abreviations like 'USA' redirects to 'The United States of America'
- Variation in Spelling like 'colour' which redirects to 'color'
- Nicknames and pseudonyms like 'Einstein' which redirects to 'Albert Einstein'

### 3.2.2. Disambiguation pages

Disambiguation pages are pages that contain the list of the different senses of a word.

### 3.2.3. Hyperlinks

Articles contain words or entities that have an article about them. So when a user clicks the link he will be redirected to an article about that word.

### 3.2.4. Category pages

Category pages are pages without articles that list members of a particular category and its subcategories. These pages have titles that start with "Category:" and is followed by the name of the particular category.

Categorization is a project of Wikipedia that attempts to assign to each article a category. The category is assigned manually by wikipedians and therefore not all pages have a category item.

Some articles belong to multiple categories. For example the article "Dominican Republic" belongs to three categories such as: "Dominican Republic", "Island countries" and "Spanish-speaking countries". Thus the article Dominican Republic appears in three different category pages.

# 4. Methodology

## 4.1. General Description

Figure 1 shows a general view of the methodology. First we extract from Wikipedia all the aligned links i.e. Wikipedia article titles. We extract the Japanese and Spanish about the same topic. Then eliminate the unnecessary data (pre-processing). Split into sentences. After extracting those articles, Use a POS tagger to add the lexical category to each word in a given sentence. Choose the sentences that match according to their lexical category. Use the dictionaries to make a word to word translation. .Finally we got the sentences to be parallel.

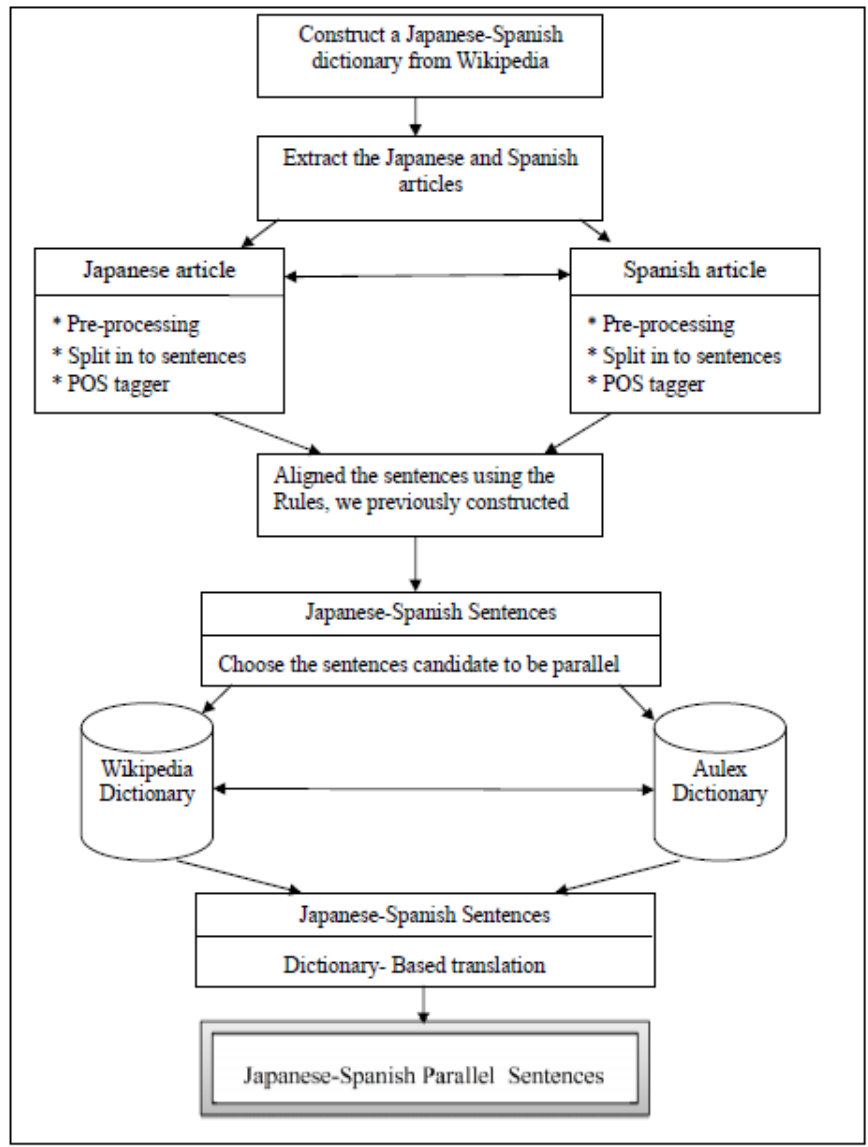

Figure 1. Methodology

## 4.2. Dictionary Extraction from Wikipedia

The goal of this phase is acquisition of Japanese-Spanish-English tuples of the Wikipedia's article titles in order to acquire translations. Wikipedia provides links in each article to corresponding articles in different languages.

Every article page in Wikipedia has on the left hand side some boxes labelled: 'navigation', 'search', 'toolbox' and at finally 'in other languages'. This has a list of all the languages available for that article, although the article in each language does not all have exactly the same contents. In most cases English articles are longer or have more information than the same article in other languages, because most of the Wikipedia collaborators are native English speakers.

### 4.2.1. Methodology

Take all articles titles that are nouns or named entities and look in the articles' contents for the box called '*In other languages*'. Verify that it has at least one link. If the box exists it redirects to the same article in other languages. Extract the words in these other languages and align it with the original article title. For instance the Spanish article titled 'economía' (economics), is translated into Japanese as 'keizaigaku' (経済学). When we click Spanish or Japanese in the other languages box we obtain an article about the same topic in the other language, this gives us the translation.

## 4.3. Extract Japanese and Spanish articles

We used the Japanese-Spanish dictionary (4.2.) to select the articles with links in Japanese and Spanish.

## 4.4. Pre-processing

We eliminate the irrelevant information from Wikipedia articles, to make processing easy and faster.
The steps are as follows.
1. Remove from the pages all irrelevant information, such as images, menus, characters such as: "()", """, "*", etc...
2. Verify if a link is a redirected article and extract the original article
3. Remove all stopwords -general words that do not give information about a specific topic such as "the", "between", "on", etc.

## 4.5. Spliting into Sentences and POS tagging

For splitting the sentences in the Spanish articles we used NLTK toolkit[5] , which is a well-known platform for building Python scripts.

For tag Spanish sentences, we used FreeLing[6], which an open source suit for language analizer, specialized in Spanish language.

---

[5] http://nltk.org/
[6] http://nlp.lsi.upc.edu/freeling/

For Splitting into sentences, in to words and add a word category, we used MeCab[7], which is a Part-of-Speech and Morphological Analyser for Japanese.

## 4.6. Constructing the Rules

Japanese is a Subject-Object-Verb language; While Spanish is a Subject-Verb-Object language.

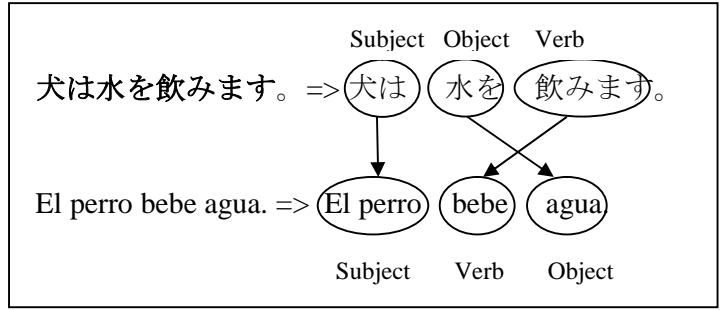

Figure 2. Basic Japanese-Spanish sentence order

Figure 2 shows the basic order of the sentences both in Japanese and Spanish, using as a example the sentence "The dog drinks water". The Japanese sentence '犬は水をのみます' (Inu wa mizu wo nomimasu.)' The dog drinks water' is translated into Spanish as 'El perro bebe agua'.

Table 1.  Japanese-Spanish rules

| Rules | Characteristics | | Rule Description |
| | Spanish | Japanese | Japanese=> Spanish |
|---|---|---|---|
| Noun | affects the gender of the adjective | Adjective do not have gender | Noun+desu => noun<br><br>Adj => Noun (gender) Adj |
| Name Entity | Always start with capital letter | Do not exist this distinction | NE=>NE (Capital letter) |
| Adjective | With gender and numbers | Adjective (Na), adjective (I) | Adj (fe/male) =>Adj (NA/I) |
| Question | It is delimited by question marks ¿? | The sentences end in か(ka) | (sentence+ か )=>( ¿ + sentence +?) |
| Pronouns | According to the context can be omitted | Can be omitted like in Spanish | Pron =>Pron |

table 1  shows some of the rules applied to this work. Those rules are created taking in account the morphological and syntactic characteristic of each language. For example, In Japanese there no exist genders for the adjectives. While in Spanish there are indispensable.

---

# 5.  Experimental Evaluation

For the evaluation of the proposed method, we took a sample of 20 random Japanese and Spanish articles. This experiments were based on two approaches: the  hyperlink approach [1] as a baseline and the Rule-Based approach. We downloaded the Wikipedia xml data for April 2012[8].

We used the Aulex[9] dictionary because the dictionary extracted from Wikipedia contain mostly noums and name entities. To align other grammatical forms such as: verbs, adjectives, etc.  we require another dictionary.

Table 2 shows the result obtained both with the baseline [1] and our approach. In column 1 shows the "Correct identification" means the sentences with the high scores and the alignment were correct. "Partial Matching" refers to the sentences both in source and target language with a noum phrase in comun. And at last "Incorrect Identification", refers to sentences with the higher scored. However, there was not even one word match

Table 2.  Results

|  | Baseline Hyperlinks | Our Approach Rule-Based POS |
|---|---|---|
| Correct Identification | 13 | 42 |
| *Partial Matching | 51 | 46 |
| Incorrect Identification | 36 | 12 |
| Total | 100 | 100 |

## 5.2. Discussion

We noticed that in the sentences form the first part of the article. It is usually the definition of the title of the article, have more correct identification, both approaches. Overall Rule-Based approach performed better than the baseline. This is because when in a given sentence, if  the hyperlink  word is or the title of the article is repeated, give automatically the best score, even if It is redundant.

Some identification performed no as well as we expected due to we need to add more rules, It can be manually or  by using bootstrapping methods, this is very interesting point for a future work. We have noticed that by using this method It is possible the construction of new sentences, even they are not in both articles.

# 6. CONCLUSIONS AND FUTURE WORKS

This study focuses on aligning Japanese-Spanish sentences by using a rule-based approach. We have demonstrated the feasibility of using Wikipedia's features for aligning several languages. We have used POS and constructed rules for aligning the sentences both in source and target article.

---

[8] The Wikipedia data is increasing constantly.
[9] http://aulex.org/ja-es/

The same method can be applied to any pair of language in Wikipedia, and another type of comparable corpora.

For future works, we will explore the used of English as a pivot language, and the automatic construction on a corpus by translating.

## ACKNOWLEDGEMENTS

We would like to thanks to Yuya R. for her contribution and helpful comments.

## Authors

Jessica C. Ramírez

She received his M.S. degree from Nara Institute of Science and Technology (NAIST) in 2007. She is currently pursuing a Ph.D. degree. Her research interest Include machine translation and word sense disambiguation.

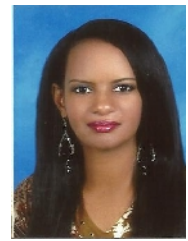

Yuji Matsumoto

He received his M.S. and Ph.D. degrees in information science from Kyoto University in 1979 and in 1989. He is currently a Professor at the Graduate School of Information Science, Nara Institute of Science and Technology. His main research interests are natural language understanding and machine learning.